\documentclass[preprint,12pt]{elsarticle}
\usepackage{amsmath,amssymb}
\usepackage{graphicx}
\usepackage{booktabs,longtable,array,calc}
\usepackage{enumitem}
\usepackage{fvextra}
\usepackage{url}
\usepackage[hidelinks]{hyperref}

\journal{Artificial Intelligence in Medicine}

\begin{document}
\begin{frontmatter}
\title{How Clinicians Think - and What AI Can Learn From It: Ordinal-First Objectives and Decision-Sufficient Abstraction for Clinical AI}
\author[aff1]{Dipayan Sengupta, MD (Dermatology)}
\author[aff2]{Saumya Panda, MD (Dermatology)}
\address[aff1]{Charnock Hospital, Kolkata, India}
\address[aff2]{Department of Dermatology, Jagannath Gupta Institute of Medical Sciences and Hospital, Kolkata, India}
\begin{abstract}
Clinical artificial intelligence is often imagined as progressing toward ever more complete representations of the patient: more variables, more modalities, more mechanistic pathways, and more precise simulation. Yet every finite clinical model is an abstraction. Before asking how accurately a model predicts, a more fundamental question arises: which distinctions in the clinical world need to be represented at all, and at what resolution? We argue that for purposive clinical decision models, this abstraction should be conditioned by the objective of the decision and constrained by the evidence available to support that level of detail. Human planning research shows that people strategically simplify task representations according to their usefulness for action, while clinical expertise research shows selective problem representation, higher-order encapsulation, and goal-directed illness and management scripts. These findings suggest that intelligence partly consists in constructing the right incomplete model rather than approximating reality uniformly. We then argue that the clinical objective that conditions this abstraction should often be ordinal-first. The primitive objects are not raw variables with fixed ranks, but objective-threshold propositions that may recur at progressively finer levels: for example, mortality below 10\% may appear early and mortality below 5\% later; treatment affordability may appear before a still lower preferred price threshold. We formalize ordered threshold refinement, define a set-valued decision relation, and introduce decision-sufficient and threshold-sufficient abstraction: a model needs only enough resolution to preserve the survivor set induced by the thresholds reached so far. Once an earlier threshold resolves the choice, further modeling of later distinctions has no decision value under the specified non-compensatory order. When a potentially decisive threshold is unresolved, additional data or model complexity should be targeted to that threshold rather than added globally. This framework unifies three claims: purpose before fidelity, order before abstraction, and refinement until decision-sufficient rather than reality-complete. It remains compatible with causal inference, mechanistic modeling, state abstraction, lexicographic optimization, constrained learning, outranking, and local cardinal refinement. The novelty is a specification doctrine for clinical AI: clinicians and patients help define what matters and how finely it must be distinguished; evidence limits what can be modeled defensibly; and AI expands, tests, simulates, and audits the resulting structure.
\end{abstract}
\end{frontmatter}

\noindent\textbf{Article type:} Position Paper

\begin{quote}
\textbf{Core proposition.} For purposive clinical AI, the objective structure is not merely downstream of prediction. It helps determine what the task-specific model should represent, at what threshold resolution, and when further refinement no longer has decision value.
\end{quote}

\section{The representation problem begins before prediction}\label{the-representation-problem-begins-before-prediction}

Clinical artificial intelligence (AI) has become increasingly capable of representing patients at high dimensionality. Models estimate risk, forecast treatment response, infer phenotypes, combine imaging with laboratory and genomic data, and simulate biological mechanisms. Yet clinical care ultimately requires actions: treat or observe, investigate or wait, admit or discharge, choose treatment A rather than treatment B.

The conventional pipeline therefore appears to be:

data -\textgreater{} representation/prediction -\textgreater{} objective function -\textgreater{} action.

The usual critique of this pipeline begins at the objective function: heterogeneous benefits, harms, burdens, and preferences are often converted into numerical contributions and combined into a scalar reward. The present thesis retains that critique but moves one step upstream.

Before an objective can be optimized, somebody has already decided what the model represents. Every finite clinical model preserves some distinctions in reality and collapses others. The first representational problem is therefore not only how precisely a model estimates its variables, but which variables, mechanisms, outcomes, and thresholds deserve explicit representation in the first place.

This matters because a model can be physiologically elaborate yet decisionally poor. A treatment simulator may resolve dozens of pharmacokinetic compartments and signaling pathways while omitting affordability, monitoring feasibility, pregnancy plans, or an intolerable adverse effect. Conversely, it may refine a two-percentage-point efficacy difference to extraordinary precision even when both regimens already exceed the clinically meaningful efficacy threshold and a later cost threshold would decide the choice.

More biological detail is not automatically more useful clinical representation. We therefore propose a two-stage role for the clinical objective. At specification time, the objective helps determine the scope and resolution of the task-specific world model.

At execution time, the same objective evaluates the predicted consequences of surviving actions. This is not circular: the stages answer different questions. First: What distinctions must be modeled?

Second: Given estimates of those distinctions, what should be done?

The architecture should not be read as a strictly one-way pipeline. During specification, the ordered objective program \(Q_{x}\) and the task-specific model \(M_{Q,E}\) are \textbf{mutually constraining}. The objective structure indicates which consequences and thresholds deserve explicit refinement, while provisional modeling, new evidence, or simulation may reveal previously unrecognized consequences that require revision of \(Q_{x}\). Specification therefore proceeds as an iterative co-refinement of objective structure and representation until both are sufficiently stable for execution.

\[K + E + x\mspace{6mu} \rightarrow \mspace{6mu} Q_{x}\mspace{6mu} \leftrightarrow \mspace{6mu} M_{Q,E}\mspace{6mu} \rightarrow \mspace{6mu} Y(a,x)\mspace{6mu} \rightarrow \mspace{6mu} Q_{x}\mspace{6mu} \rightarrow \mspace{6mu} D_{Q}(x).\]

The central principles are therefore: \textbf{purpose before fidelity; order before abstraction; and refine until decision-sufficient, not until reality-complete.}

\section{Every clinical model is a selective abstraction}\label{every-clinical-model-is-a-selective-abstraction}

\subsection{Open-ended reality and finite representation}\label{open-ended-reality-and-finite-representation}

Clinical reality is open-ended relative to any finite computational representation. Biology spans molecular, cellular, tissue, organ, behavioral, social, and environmental scales; observations are incomplete; mechanisms remain uncertain; and clinically important context may be difficult to measure or to encode. This does not require the metaphysical claim that nature is literally infinite in complexity.

It is enough that no practical clinical model can preserve every potentially relevant distinction. Modeling theory has long treated simplification as unavoidable rather than defective. Levins emphasized that useful biological models trade among competing desiderata and are built for purposes rather than as miniature replicas of reality \cite{ref1}.

Modern digital-twin theory reaches a compatible conclusion: The National Academies describes virtual representations as needing to be fit for purpose, with model type, fidelity, resolution, parameterization, and quantities of interest selected - and potentially adapted - for the decision task and available computational resources \cite{ref2}. The relevant question is therefore not whether a model is complete, but whether the omissions are safe for its intended use. Let \(W\) denote the clinically relevant world and let \(\phi\) map that world into a computational state space \(Z\): \(\phi:W \rightarrow Z\). Because \(\phi\) is necessarily many-to-one, different states of the world will share the same modeled representation.

The scientific problem is to decide which distinctions may be collapsed without changing the intended decision.

\subsection{Complexity must also be supported by evidence}\label{complexity-must-also-be-supported-by-evidence}

A second limit is epistemic rather than computational. Even when a biological mechanism is real, available observations may not support estimating it at the desired resolution. Dynamic biological models can contain many unknown or immeasurable parameters; calibration and identifiability problems tend to worsen as model size and the number of unknowns increase \cite{ref3}.

Model-reduction work similarly emphasizes matching model complexity to the information content of the data rather than assuming that additional detail is automatically informative \cite{ref4}. The literature on sloppy multiparameter models makes the point more sharply. Complex mechanistic systems often exhibit behavior controlled by a relatively small number of stiff parameter combinations while many other combinations are poorly constrained and have little influence on predictions of interest \cite{ref5,ref6}.

Thus, more mechanistic detail does not necessarily produce more patient-specific knowledge. It can instead increase the number of assumptions, priors, and unidentifiable quantities hidden inside an apparently precise simulation. This yields a supporting constraint for clinical AI:

\textbf{Justified model detail = decision relevance AND evidential support.}

Decision relevance is the primary concern of this thesis; evidential support is the boundary that prevents relevant but weakly supported distinctions from being represented with false precision.

\section{How humans make complex worlds tractable}\label{how-humans-make-complex-worlds-tractable}

\subsection{Goal-conditioned abstraction in human cognition}\label{goal-conditioned-abstraction-in-human-cognition}

Human cognition suggests one way an intelligent system can act in a world too complex to represent exhaustively. Ho and colleagues challenged the assumption that planning begins from a fixed, complete task representation. In preregistered behavioral experiments, people strategically simplified their mental representations and balanced representational complexity against usefulness for planning and action \cite{ref7}.

Resource-rational theories provide a broader normative account: limited memory, time, attention, and computation should be allocated according to their value for achieving the task rather than toward maximal internal fidelity \cite{ref8}. A goal-centric account of learning goes further. Molinaro and Collins review evidence that goals influence the representation of inputs, states, actions, and outcomes themselves \cite{ref9}.

The objective therefore need not enter only after a world model has been constructed. It can shape which distinctions are represented in that world model.

\textbf{Model-first view:} world -\textgreater{} fixed representation -\textgreater{} optimize goal\\
\textbf{Purpose-conditioned view:} goal -\textgreater{} goal-relevant representation -\textgreater{} planning/action

This does not imply that humans are always optimal, unbiased, or consciously aware of their representational strategy. It supports a narrower proposition: effective intelligence can arise by constructing a task-appropriate incomplete representation rather than carrying a uniformly detailed model of the environment.

\subsection{Clinical expertise shows compatible compression}\label{clinical-expertise-shows-compatible-compression}

Clinical reasoning research shows a closely related pattern. Expertise is associated with multiple coordinated knowledge representations rather than one universal reasoning algorithm \cite{ref10}. With experience, detailed biomedical knowledge becomes encapsulated into higher-order concepts and illness scripts that can be activated without reconstructing every mechanistic step explicitly \cite{ref11}.

Script theory describes these structures as goal-directed knowledge organizations adapted to performing tasks efficiently; they generate expectations, guide information search, support inference, and organize action \cite{ref12}. Recent direct evidence makes this compression visible. In a comparison of internal-medicine faculty and early residents, expert physicians included fewer irrelevant comorbidities and more often encapsulated clinical data into higher-order terms; well-structured problem representations were also associated with diagnostic accuracy \cite{ref13}.

In management reasoning, empirically derived management scripts connect options, clinician tasks, patient preferences, values, constraints, education needs, interactions, and encounter flow \cite{ref14,ref15}. These studies do not prove that clinicians consciously construct a mathematical ordered list before every decision. They establish the empirical bridge we need: expert clinical cognition uses selective, compressed, context-sensitive and task-sensitive representations.

The proposed ordinal architecture is a normative theory for how such relevance can be specified and audited computationally.

\section{From broad clinical knowledge to a decision-specific world model}\label{from-broad-clinical-knowledge-to-a-decision-specific-world-model}

\subsection{Background knowledge is not the task-specific model}\label{background-knowledge-is-not-the-task-specific-model}

Clinicians do not rebuild physiology on the fly. They enter a decision with broad domain knowledge \(K\): mechanistic knowledge, trial evidence, guideline constraints, learned patterns, prior cases, and experience with feasibility. The patient and care environment add context \(x\), including preferences, resources, goals, and practical constraints.

This broad knowledge is much larger than the representation needed for one choice. The act of clinical reasoning therefore involves selection. A generic high-level concern such as minimizing serious harm can be instantiated into the particular risks relevant to the current therapies; evidence and context determine which risks are plausible enough to model, and clinical judgment determines how they should be ordered relative to efficacy, burden, feasibility, cost, and patient priorities.

The key claim is not that preference precedes all knowledge. It is that, once a purposive clinical decision is defined, the objective structure helps select which parts of broad knowledge require explicit quantitative representation for that task.

\subsection{The order is over threshold propositions, not raw variables}\label{the-order-is-over-threshold-propositions-not-raw-variables}

The primitive objects in this framework are objective-threshold propositions, not raw dimensions with one permanent rank. The same outcome can recur at multiple places in the order at progressively finer thresholds. For example, treatment cost may appear early as a feasibility condition and much later as a preference refinement:

\begin{itemize}
\item
  \(q_{3}\): monthly cost \textless{} INR 1000 (affordable enough to be feasible)
\item
  \(q_{17}\): monthly cost \textless{} INR 100 (preferred if earlier objectives remain tied)
\end{itemize}

The first threshold can be important for decision even though exact price differences below INR 1000 are initially irrelevant.

If both treatments pass \(q_{3}\), a model need not distinguish INR 500 from INR 800 at that stage. If an earlier safety or efficacy threshold later resolves the choice, finer price modeling is unnecessary. If the decision survives to \(q_{17}\), the same dimension becomes active again at finer resolution.

Mortality, pain, toxicity, function, monitoring burden, and other dimensions can be treated similarly. This is why ordinal-first does not mean assigning a small weight to a low-ranked variable. It means giving a threshold conditional decisional authority: irrelevant while an earlier threshold discriminates, potentially decisive when its turn is reached.

\subsection{Threshold-sufficient abstraction}\label{threshold-sufficient-abstraction}

Repeated thresholds imply a new representational principle. A clinical model need not preserve the full cardinal state of a variable when the current decision only requires classification relative to a threshold. Representation can become progressively finer only when the decision reaches a threshold that requires finer discrimination.

Call an abstraction threshold-sufficient through criterion \(k\) if it preserves the survivor set produced by evaluating the first \(k\) threshold propositions. Let \(S_{k}(w)\) be the set of actions still admissible after criteria 1 through \(k\) in world state \(w\). Then a conceptual sufficiency condition is:

\[\phi_{k}\left( w_{1} \right) = \phi_{k}\left( w_{2} \right) \Rightarrow S_{k}\left( w_{1} \right) = S_{k}\left( w_{2} \right).\]

If \(\left| S_{k} \right| = 1\), later criteria do not need to be represented for that choice under the specified non-compensatory order. If multiple actions survive, the representation is refined only as needed to evaluate the next potentially discriminating threshold. Thus, threshold refinement can induce representational refinement.

This sufficiency claim concerns \textbf{decision-directed resolution}, not every internal variable required for prediction. A mechanistic, latent, or contextual variable may still need to be represented if it is necessary to estimate an earlier high-priority consequence accurately, even when that variable is not itself an objective in \(Q_{x}\). Thus, ``do not refine an unreached objective'' means that the system should not increase resolution \emph{solely} to distinguish alternatives on that unreached objective; it does not license discarding information needed to estimate thresholds that have already been reached.

\begin{figure}[htbp]
\centering
\includegraphics[width=\textwidth]{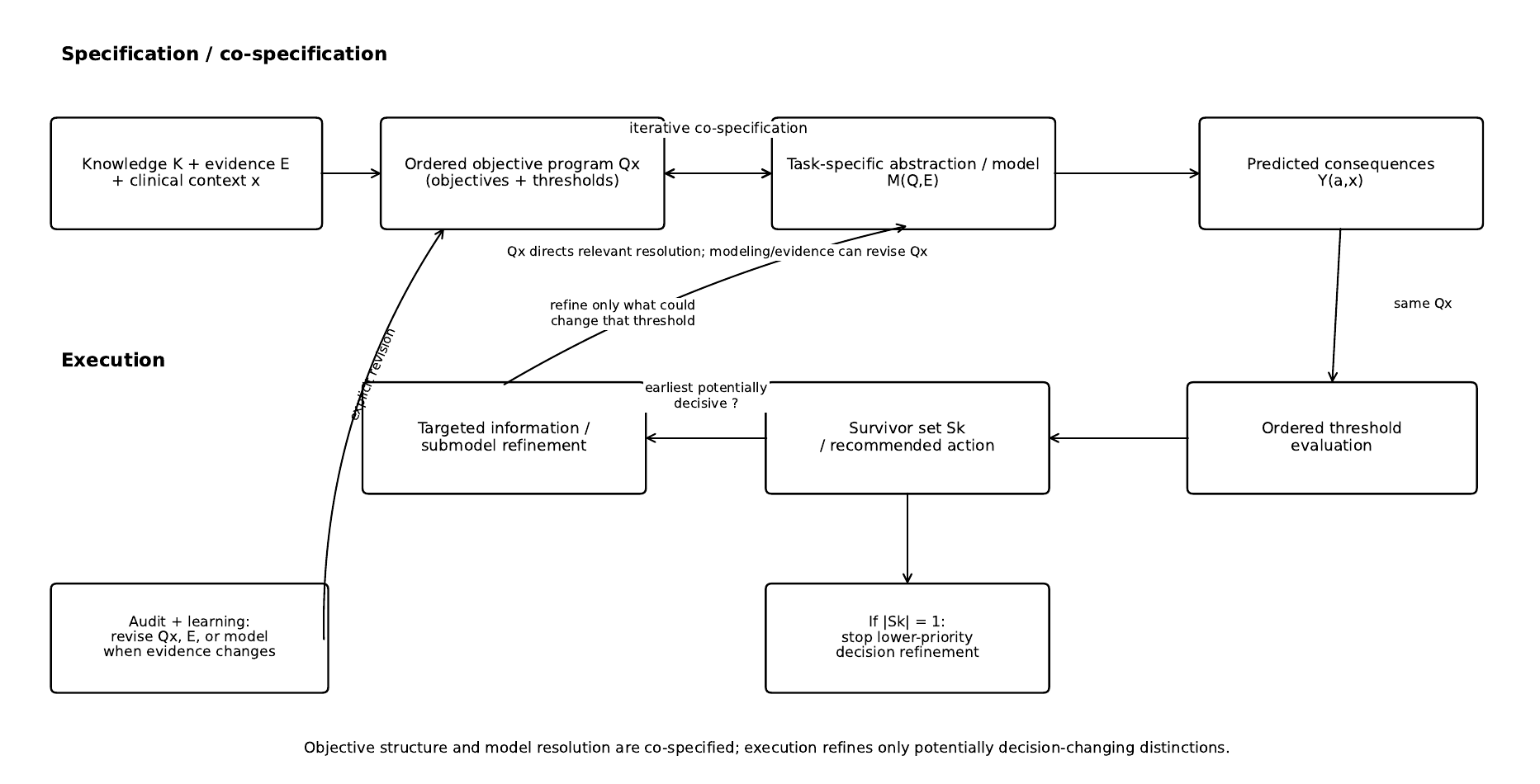}
\caption{Iterative specification and execution in ordinally conditioned clinical modeling. The objective program and task-specific model are co-specified: objectives determine which distinctions deserve refinement, while modeling and evidence can reveal consequences that revise the objective program. During execution, unresolved potentially decisive thresholds trigger targeted refinement; once a survivor set is resolved, lower-priority decision-directed refinement stops.}
\label{fig:architecture}
\end{figure}

\section{Why the clinical objective should be ordinal-first}\label{why-the-clinical-objective-should-be-ordinal-first}

Once the objective is recognized as upstream of task-specific modeling, the nature of that objective becomes more consequential. A common answer is a scalar reward U that embeds all relevant consequences into one compensatory numerical structure. This may be justified when the clinical problem genuinely supports stable, normatively acceptable cardinal trade-offs.

It should not be presumed by default. Classical measurement theory separates order from stronger notions of distance. Ordinal claims survive monotone transformations, whereas interval and ratio interpretations require stronger invariance assumptions \cite{ref16,ref17}.

Lexicographic preference theory separately demonstrates that coherent priority relations need not collapse into ordinary additive utility \cite{ref18,ref19}. Expected-utility and multiattribute-utility theory remain legitimate when their axioms and elicitation assumptions are satisfied; what we object is unearned cardinalization. Human preference is frequently elicited through comparison.

Health-economic methods include ranking and discrete-choice approaches that infer latent preference structure from comparative choices rather than observing a psychological ruler directly \cite{ref20,ref21}. Preference-sensitive decisions can also involve preference construction during deliberation rather than retrieval of a fixed internal number \cite{ref22,ref23}. The consequences of clinical actions are heterogeneous: survival, irreversible disability, pain, itch, uncertainty, monitoring burden, stigma, work, caregiving, cost, and outcomes the patient may never have experienced.

Measurement of a consequence is therefore not identical to valuation of that consequence. Before assigning numerical exchange rates among heterogeneous objectives, a decision architecture should at least establish which comparisons matter, what thresholds count as sufficiently satisfied, and which lower-priority gains are allowed to compensate for earlier distinctions. The central principle remains:

\textbf{Order before distance. Threshold before trade-off.}

The revised thesis adds two upstream corollaries:

\textbf{Purpose before fidelity. Order before abstraction.}

\section{Ordered threshold refinement}\label{ordered-threshold-refinement}

\subsection{Why strict lexicography is too rigid}\label{why-strict-lexicography-is-too-rigid}

Pure lexicographic optimization is brittle when any nonzero difference on an earlier objective dominates arbitrarily large differences later. Clinical reasoning is usually less rigid because priorities are expressed through acceptable regions and critical values rather than infinitesimal differences.

\begin{longtable}[]{@{}ll@{}}
\toprule\noalign{}
\textbf{Priority} & \textbf{Objective-threshold proposition} \\
\midrule\noalign{}
\endhead
\bottomrule\noalign{}
\endlastfoot
1 & Probability of death \textless{} 10\% \\
2 & Probability of paralysis \textless{} 30\% \\
3 & Probability of death \textless{} 5\% \\
4 & Probability of severe irreversible toxicity \textless{} 10\% \\
5 & Clinically meaningful functional recovery achieved \\
6 & Severe pain avoided \\
7 & Severe itch avoided \\
8 & Monitoring burden acceptable \\
9 & Treatment affordable \\
10 & Absolute price below a lower preferred threshold \\
\end{longtable}

The same dimension can therefore appear repeatedly: mortality \textless10\% is an early safety threshold, while mortality \textless5\% is a later refinement. A low-priority price threshold is not a tiny coefficient. It contributes nothing while an earlier criterion discriminates and can decide the case completely when all earlier criteria are tied.

\subsection{Decision-theoretic and clinical precedents}\label{decision-theoretic-and-clinical-precedents}

Ordered reasons, aspiration levels and stopping rules are established decision-theoretic objects. The priority heuristic evaluates reasons sequentially and stops when a critical difference is reached, while lexicographic semiorders generalize thresholded priority relations \cite{ref24,ref25}. The priority heuristic is not a universal descriptive model - compensatory integration also occurs - but its structure demonstrates that priority and threshold are distinct from finite weighting \cite{ref26}.

Medicine has an independent threshold tradition. Pauker and Kassirer formalized test and treatment thresholds that convert continuous probabilities into action regions \cite{ref27}; later work linked threshold models with signal detection, evidence accumulation and fast-and-frugal decision trees \cite{ref28}. Direct physician studies have fitted ordered critical-value heuristics to clinical judgments, including pediatric asthma admission decisions \cite{ref29}, and fast-and-frugal models have approximated some medical judgments with fewer cues than regression alternatives \cite{ref30}.

Cognitive-task analysis has also extracted simple if-then heuristic structures from complex clinical work \cite{ref31}. The same structural family has been used in lung-cancer decision pathways, pharmacogenomic medication prioritization, and reformulation of clinical guidelines into fast-and-frugal trees \cite{ref32,ref33,ref34}. These studies do not show that expert clinicians always reason lexicographically.

They support the narrower proposition that clinical judgment often contains ordered cues, critical values, aspiration levels, and stopping rules that unrestricted additive compensation does not faithfully represent.

\section{Formal specification of an ordinally conditioned clinical model}\label{formal-specification-of-an-ordinally-conditioned-clinical-model}

\subsection{Actions, evidence and consequences}\label{actions-evidence-and-consequences}

Let \(A = \{ a_{1},\ldots,a_{n}\}\) be candidate actions in clinical context \(x\). Let \(K\) denote broad domain knowledge and \(E\) the evidence available to support the current decision. When actions alter outcomes, consequence estimates should be informed by randomized evidence, causal estimates, mechanistic knowledge, or other defensible intervention evidence rather than by prognosis under observed care alone \cite{ref35,ref36}.

Define the ordered program as

\[Q_{x} = \left( q_{1},q_{2},\ldots,q_{m} \right),\]

with objective-threshold propositions

\[q_{k}(a,x) = 1\{ Y_{j_{k}}(a,x) \in T_{k}\}.\]

The outcome index \(j_{k}\) may repeat, allowing the same outcome to recur at progressively finer thresholds. The order \(q_{1} \succ q_{2} \succ \cdots \succ q_{m}\) means ``consider earlier when discriminating among the current alternatives''; it is not an intrinsic ranking of treatments or raw variables.

\subsection{The objective conditions the task-specific model}\label{the-objective-conditions-the-task-specific-model}

Instead of assuming a fixed representation layer that estimates every potentially available consequence, define a task-specific abstraction \(\phi_{Q,E}\) and model \(M_{Q,E}\). The ordered program identifies the distinctions that may become decisionally relevant; the evidence \(E\) limits the resolution at which those distinctions can defensibly be estimated. The conceptual world \(W\) is not available in full. The notation simply makes explicit that the model collapses many real states into a smaller representation.

The aim is not to minimize representational error everywhere but to preserve distinctions that can alter the clinical decision.

\[\phi_{Q,E}:W \rightarrow Z_{Q,E}.\]

The co-specification loop is therefore bidirectional: \(Q_{x}\) tells the model where fidelity is decisionally valuable, while \(M_{Q,E}\) and new evidence can expose consequences, interactions, or uncertainties that require revision of \(Q_{x}\). This loop should stabilize before a decision is executed, but it remains revisable as evidence or context changes.

\subsection{Set-valued decision relation and decision sufficiency}\label{set-valued-decision-relation-and-decision-sufficiency}

Clinical reasoning need not force a unique action. Let \(D_{Q}(w)\) be the set of actions that remain admissible under the ordered program in world state \(w\): \(D_{Q}(w) \subseteq A\).

This resembles policy-preserving state abstraction in reinforcement learning, where states may be aggregated when distinctions are irrelevant to the policy or value structure \cite{ref37}. Bisimulation-based methods similarly formalize behavioral similarity and state aggregation for Markov decision processes \cite{ref38}. Our contribution is not a new state-abstraction theorem.

It is the clinical specification of what must be preserved: a context-sensitive ordered set of threshold propositions grounded in evidence, patient values, and clinical priorities.

A conceptual abstraction is decision-sufficient with respect to \(Q\) when

\[\phi\left( w_{1} \right) = \phi\left( w_{2} \right) \Rightarrow D_{Q}\left( w_{1} \right) = D_{Q}\left( w_{2} \right).\]

This exact definition is deliberately idealized. In clinical deployment, a natural extension is \textbf{probabilistic or robust decision sufficiency}, in which the residual probability that distinctions collapsed by an abstraction would alter the survivor set is explicitly bounded rather than assumed to be zero.

\subsection{Sequential set reduction}\label{sequential-set-reduction}

For multiple actions, let \(S_{0} = A\). When criterion \(k\) is resolved for all actions in \(S_{k - 1}\), retain the actions with the best threshold status. With binary pass/fail criteria:

\[S_{k} = \{ a \in S_{k - 1}:q_{k}(a,x) = \max_{b \in S_{k - 1}}q_{k}(b,x)\}.\]

If \(\left| S_{k} \right| = 1\), the choice is resolved and later criteria are not reached. If several actions survive, continue. If all relevant ordinal criteria tie, a set-valued recommendation may be returned, or a local cardinal refinement may be applied when its assumptions are defensible.

\subsection{Uncertainty and targeted refinement}\label{uncertainty-and-targeted-refinement}

Clinical estimates are uncertain. Replace binary threshold status with a three-valued state. Uncertainty in health care has multiple forms and may warrant different responses \cite{ref39}.

A potentially decisive ``?'' should pause ordered reduction. The system can acquire additional data, ask a preference question, obtain expert review, refine the relevant submodel, issue a conditional recommendation, or - if delay itself carries risk - invoke an explicitly specified act-now-versus-wait policy. Lower-priority gains should not silently convert the unresolved high-priority question into a soft penalty.

This creates adaptive abstraction. Model complexity is increased locally when additional resolution could change the status of the earliest potentially discriminating threshold. If the threshold is resolved and leaves one action, later objectives do not require further simulation for that choice.

\[q_{k}(a,x) \in \{ 1,0,?\},\]

where ``?'' means that current evidence or model resolution does not establish whether the target is met. A potentially decisive ``?'' pauses ordered reduction at criterion \(k\); it is not ranked between 0 and 1. Lower-priority criteria cannot resolve the choice until the uncertainty is resolved or an explicit act-under-uncertainty policy is invoked.

\subsection{Cardinal refinement remains available}\label{cardinal-refinement-remains-available}

The architecture is not ordinal instead of cardinal. True ties and genuinely tradeable domains remain. If all relevant ordered criteria are tied for a surviving set \(S_{m}\), a local cardinal model may be used, provided the required measurement and preference assumptions are defensible.

Expected utility, multiattribute utility, cost-effectiveness, minimax regret, and learned preference models can all occupy this local role \cite{ref40,ref41}. Cardinal information is used where it earns interpretability rather than being presumed globally.

\[a^{*} \in arg\max_{a \in S_{m}}U_{local}(a \mid x).\]

\section{Ordinal fidelity and compensatory leakage}\label{ordinal-fidelity-and-compensatory-leakage}

\subsection{Ordinal fidelity}\label{ordinal-fidelity}

A decision rule has ordinal fidelity to \(Q_{x}\) when its recommendation respects the first resolved threshold that distinguishes the surviving alternatives. For a pair a,b, if the first resolved discriminating criterion favors a, a faithful decision rule must not select b unless the order itself is explicitly revised. The reference is not historical clinician behavior.

Historical choices are evidence about structure, not ground truth. The reference is the currently justified ordered program, which may be challenged by better evidence, patient values, or new context.

\subsection{Compensatory leakage}\label{compensatory-leakage}

We call an unauthorized reversal compensatory leakage: lower-priority gains overturn a decision already determined by an earlier criterion without explicit revision of that ordering. A scalar model can exhibit compensatory leakage even when every component prediction is statistically accurate, because the error lies in the semantics of combination rather than the calibration of individual outcomes. A useful stress test holds an early threshold distinction fixed while progressively improving lower-priority features of the losing action.

If an unconstrained score eventually flips the recommendation, the model has introduced a compensation not authorized by \(Q_{x}\). Across a test suite, this can be summarized as a priority-violation rate.

\section{Why decision-sufficient abstraction can be robust under incomplete context}\label{why-decision-sufficient-abstraction-can-be-robust-under-incomplete-context}

We use robust in a limited representational sense. We do not claim that ordinal rules dominate expected utility in every well-specified decision problem. We claim that purpose-conditioned threshold representations can make fewer unsupported assumptions when the decision-relevant world model is incomplete.

Clinical context is extraordinarily heterogeneous. A recent scoping review catalogued hundreds of distinct contextual influences across patients, clinicians, families, institutions, colleagues, diseases, and treatments \cite{ref42}. Situated-cognition research likewise emphasizes that clinical reasoning performance depends on interactions among the patient, clinician, task, and environment \cite{ref43,ref44}, while some experience-based expertise remains difficult to exhaustively articulate \cite{ref45}.

A global scalar model is vulnerable in two ways. First, it may omit a dimension important for decision. Second, it may impose numerical exchange rates among dimensions whose trade-offs are unstable, context-sensitive, or never actually elicited.

Ordered thresholds reduce one route for error by requiring fewer global trade-off assumptions and by preventing many low-priority gains from accumulating against an earlier non-compensatory distinction. The new abstraction argument adds a third protection: irrelevant precision can be avoided. If a threshold already classifies all surviving actions identically, finer representation of that dimension is unnecessary until a later, stricter threshold is reached.

If an earlier criterion resolves the choice, unreached dimensions need not be simulated further for that decision. This is a representational stopping rule, not only a decision stopping rule. None of this makes omissions harmless.

If a factor would change an early threshold but is absent from K, E, Q\_x, or the predictive model, the system can still fail. The advantage is diagnostic transparency: the failure can be localized as missing knowledge, missing evidence, missing objective, inadequate threshold, or inadequate model resolution rather than disappearing inside an opaque coefficient vector.

\section{Collaboration in the specification loop}\label{collaboration-in-the-specification-loop}

The hardest problem is constructing \(Q_{x}\) and the associated abstraction. Conventional human-in-the-loop designs often place clinicians after the model output: the AI predicts, the clinician reviews, and the clinician accepts or overrides. Ordinally conditioned modeling requires human contribution further upstream.

The ordered program and task-specific model should draw on several sources:

\begin{itemize}
\item
  causal and trial evidence about consequences of actions;
\item
  mechanistic and pharmacologic knowledge relevant to plausible harms and benefits;
\item
  clinical guidelines, safety rules, and governance constraints;
\item
  expert clinical reasoning and case-specific context;
\item
  patient preferences, lived experience, feasibility, resources, and tolerability;
\item
  real-world outcomes and treatment heterogeneity; and
\item
  AI-assisted search for missing objectives, interactions, evidence, and inconsistencies. This collaboration can occur at three levels. At design time, domain experts and patient representatives define generic objective families, important thresholds, provenance, and known precedence relations.
\end{itemize}

At case time, clinicians and patients instantiate or reorder them for the individual context. After decisions, outcomes, overrides, disagreements, and new evidence test whether the order or abstraction should be revised. Shared decision making therefore becomes part of constructing the objective structure rather than consent appended after an optimizer has finished.

Established shared-decision models and values-clarification methods already treat preferences as something that may require explicit deliberation \cite{ref46,ref47,ref48}. The clinician contribution should not be romanticized. Clinical judgment can be biased or wrong, and automation can induce inappropriate reliance on incorrect decision support \cite{ref49}.

AI should learn the representational discipline of ordered discrimination while remaining able to challenge the content using broader evidence and explicit provenance. Human oversight is strongest when normative choices are inspectable, not when humans merely approve a finished score. For regulated systems, logging the threshold that determined a recommendation, the evidence used to estimate it, which criteria were not reached, and any explicit revision to the order is directionally aligned with contemporary expectations for traceable human oversight and controlled model change \cite{ref50,ref51}.

\section{Relation to existing AI and decision paradigms}\label{relation-to-existing-ai-and-decision-paradigms}

\subsection{State abstraction and bisimulation}\label{state-abstraction-and-bisimulation}

State abstraction in planning and reinforcement learning explicitly ignores state information that is irrelevant to finding an effective policy, allowing solutions to be found in a more compact abstract state space \cite{ref37}. Bisimulation and related metrics formalize when states can be treated as behaviorally similar for decision processes \cite{ref38}. Decision-sufficient and threshold-sufficient abstraction belong to this broad computational family.

The clinical addition is upstream normative specification. In an MDP, the reward and transition structure are usually assumed to be given. In clinical AI, the difficult problem is precisely what the reward-equivalent structure should be, which thresholds constitute meaningful equivalence classes, and whose values determine them.

The ordered program supplies those semantics before selecting an implementation.

\subsection{Lexicographic optimization and safe reinforcement learning}\label{lexicographic-optimization-and-safe-reinforcement-learning}

The execution rule is closely related to lexicographic optimization, lexicographic semiorders, and lexicographic multi-objective reinforcement learning \cite{ref18,ref25,ref52}. Safe reinforcement learning provides a mature engineering literature on statewise, probabilistic, expected, and other constraint types \cite{ref53,ref54}. These are compatible implementation tools rather than the conceptual novelty claimed here.

A statewise prohibition, a chance constraint, an expected cumulative cost bound, and a finite penalty are not semantically interchangeable. If a clinical priority is implemented only as a soft penalty, sufficiently many gains elsewhere can overturn it. Ordinal-first design therefore specifies the intended decision relation first and then chooses a computational mechanism that faithfully implements it.

\subsection{Outranking, screening and hybrid symbolic-statistical systems}\label{outranking-screening-and-hybrid-symbolic-statistical-systems}

Outranking methods such as ELECTRE represent preference, indifference, and veto-like relations without one globally compensatory score, while the broader multi-criteria literature provides many technical variants \cite{ref55,ref56}. Elimination by aspects and screening-plus-compensation models similarly show that sequential elimination and later trade-off can coexist \cite{ref57,ref58}. Neuro-symbolic clinical systems provide an implementation pattern in which learned representations coexist with explicit auditable structure \cite{ref59}.

\subsection{Fit-for-purpose simulation and digital twins}\label{fit-for-purpose-simulation-and-digital-twins}

Digital twins are a useful but nonexclusive illustration. A patient-specific simulation can contain PDEs, ODEs, Bayesian state estimation, and learned surrogates while still be decisionally incomplete if it resolves the wrong distinctions. Conversely, a deliberately uneven model - highly detailed in one physiological subsystem, categorical in another, and exact about a practical constraint such as price - may be better suited to the decision.

This aligns with the National Academies' fit-for-purpose view that model fidelity, resolution, parameterization, and quantities of interest should be selected and potentially adapted to the particular decision task \cite{ref2}.

\section{Clinical illustrations}\label{clinical-illustrations}

\subsection{Affordability as repeated thresholds}\label{affordability-as-repeated-thresholds}

Consider two regimens with similar clinically meaningful efficacy and acceptable major safety. A high-priority feasibility proposition might be monthly cost \textless{} INR 1000. Treatment A costs INR 600 and treatment B INR 800; both pass, so the exact INR 200 difference need not influence the decision yet.

If an earlier efficacy or toxicity threshold later distinguishes the actions, price below INR 1000 has done all the representational work required and finer price simulation is unnecessary. If the actions survive to a later preference threshold such as monthly cost \textless{} INR 700, the same variable becomes relevant at finer resolution and A now passes while B fails. The example illustrates why the order is over threshold propositions, not over variables: cost can be high-priority as feasibility and low-priority as preference refinement without contradiction.

\subsection{Teratogenic treatment: contextual priority rather than metaphysical veto}\label{teratogenic-treatment-contextual-priority-rather-than-metaphysical-veto}

For severe acne, isotretinoin is contraindicated in pregnancy because of embryo-fetal toxicity \cite{ref60}. In that ordinary context, a threshold such as no current pregnancy appears very early; superior acne efficacy does not rescue the option when the threshold is not met. But a hypothetical life-threatening maternal condition with only a teratogenic lifesaving therapy would create a different order in which immediate maternal survival may precede fetal-risk minimization.

What appears to be an absolute veto is better represented as a high-priority contextual threshold embedded in a broader order.

\subsection{Pain and itch: repeated severity thresholds and local cardinality}\label{pain-and-itch-repeated-severity-thresholds-and-local-cardinality}

A plausible program for a patient who cares about both pain and itch might include: avoid severe pain, avoid severe itch, avoid moderate pain, avoid moderate itch, then refine mild residual symptoms. This does not mean pain must be eliminated before itch is considered. Severity and symptom type jointly define ordered thresholds.

If two treatments both avoid severe pain and severe itch, lower thresholds become active. If validated measurement and preference elicitation support meaningful distances at that point, local cardinalization can refine the choice; stored numbers are not automatically a common currency.

\subsection{Active bleeding and anticoagulation}\label{active-bleeding-and-anticoagulation}

In atrial fibrillation with major active bleeding, immediate hemorrhage control can precede finer comparison of long-term stroke prevention; after bleeding is controlled, stroke prevention again becomes central \cite{ref61}. The logic differs from assigning active bleeding a large negative coefficient. A sufficiently large predicted stroke-prevention benefit should not silently compensate for uncontrolled life-threatening bleeding at that decision point unless the order is explicitly revised.

\section{Evaluation: test the representation, not only the predictor}\label{evaluation-test-the-representation-not-only-the-predictor}

Prediction-layer metrics remain necessary: discrimination, calibration, uncertainty, external validity, and causal validity when actions are compared. Decision-curve analysis and related tools remain useful for evaluating action- level clinical utility \cite{ref62,ref63}. But an ordinally conditioned architecture creates additional requirements.

\begin{longtable}[]{@{}
  >{\raggedright\arraybackslash}p{(\columnwidth - 2\tabcolsep) * \real{0.2200}}
  >{\raggedright\arraybackslash}p{(\columnwidth - 2\tabcolsep) * \real{0.7800}}@{}}
\toprule\noalign{}
\begin{minipage}[b]{\linewidth}\raggedright
\textbf{Evaluation target}
\end{minipage} & \begin{minipage}[b]{\linewidth}\raggedright
\textbf{Question}
\end{minipage} \\
\midrule\noalign{}
\endhead
\bottomrule\noalign{}
\endlastfoot
Ordinal fidelity & Does the recommendation respect the first resolved discriminating threshold? \\
Compensatory leakage & Can accumulated lower-priority gains reverse an earlier authorized distinction? \\
Threshold sufficiency & Does the current abstraction preserve the survivor set through the thresholds reached so far? \\
Refinement efficiency & When a threshold is unresolved, does added information/modeling target the potentially decisive distinction rather than expand globally? \\
Ordering coverage & Are clinically important objectives and relevant threshold states represented? \\
Evidence support & Are patient-specific parameters and mechanisms constrained by evidence at the resolution claimed? \\
Boundary behavior & Does an unresolved early threshold trigger testing, refinement, conditional advice, or an explicit act-under-uncertainty policy? \\
Human factors & Can clinicians identify what threshold determined the action, what was not modeled further, and what evidence would change the plan? \\
Patient concordance & Does the order reflect patient priorities where appropriate, and how burdensome is preference elicitation? \\
\end{longtable}

DECIDE-AI provides a framework for early-stage clinical evaluation of AI decision support, including workflow and human factors \cite{ref64}. The revised theory adds a specific falsifiable prediction: after an earlier threshold discriminates, perturbations confined to unreached lower-priority objectives should not change the recommendation. A second prediction is representational: additional fidelity confined to unreached objectives should not improve the decision under the specified program.

\section{Research agenda}\label{research-agenda}

The scientific question is no longer only `what are the correct weights?' It becomes: \textbf{what ordered threshold structure best represents a clinically sound decision, what abstraction is sufficient to evaluate that structure, and how should both be learned, personalized, and revised?}

\begin{itemize}
\item
  Decision decomposition studies should elicit clinician threshold sequences using vignettes, cognitive interviews, process tracing, observed decisions, and cognitive-task analysis, comparing ordered and compensatory models rather than assuming either in advance \cite{ref29,ref31}.
\item
  Clinical abstraction studies should test whether experts selectively request, represent, or simulate information according to the earliest potentially discriminating objective, extending existing evidence on expert problem representation \cite{ref13}.
\item
  Patient preference studies should compare ordered thresholds, compensatory utilities, and hybrid screening-plus-compensation strategies, especially for heterogeneous outcomes that are difficult to place on one stable numerical scale.
\item
  Learning methods should infer candidate objective-threshold programs from guidelines, evidence, clinician reasoning, patient preferences, and outcomes while preserving provenance. Learning one scalar reward is not equivalent to learning an ordered decision representation.
\item
  Adaptive modeling methods should maintain model libraries at multiple resolutions and refine only the submodels needed to classify the current threshold. The comparison should be against globally high-fidelity approaches on computational burden, evidence requirements, safety, and action-level outcomes.
\item
  Evidence-overreach studies should quantify when added mechanistic detail improves patient-specific prediction versus merely increasing dependence on priors or non-identifiable parameters \cite{ref3,ref4}.
\item
  Formal work should develop \textbf{probabilistic and robust decision-sufficiency criteria}, for example by bounding the probability that distinctions collapsed by an abstraction would change the surviving action set above an acceptable tolerance \(\epsilon\).
\item
  Prospective evaluations should compare score-first, fixed-model ordinal, and adaptive ordinal-abstraction systems on safety, utility, clinician comprehension, patient concordance, compensatory leakage, and model-refinement burden.
\item
  AI should be evaluated not only on imitation of historical clinicians but on whether it helps construct better orderings and abstractions: more evidence-responsive, more patient-sensitive, less biased, and more consistent with long-term outcomes.
\end{itemize}

\section{Boundary conditions}\label{boundary-conditions}

Ordinal-first is a default representational discipline for purposive clinical decision models, not a universal theory of scientific modeling. A basic-science model may legitimately represent mechanisms whose decision relevance is unknown because discovery itself is the goal. Once such a model is used to choose an intervention, however, an objective again becomes necessary to determine which predictions matter for that decision.

Nor is the proposal a universal ban on scalarization. If a clinical problem has a well-validated, stable, and normatively acceptable cardinal utility over all relevant consequences, global expected-utility optimization may be appropriate. The framework challenges unearned cardinalization, not cardinal decision theory.

Thresholds and orderings are not value-free. Choosing mortality \textless10\% rather than \textless8\%, deciding whether short-term relief precedes treatment burden, and deciding whose preference has authority are substantive judgments. The advantage is that these judgments are exposed and auditable rather than distributed invisibly across coefficients.

The order can be wrong. Diagnostic-error research documents cognitive biases and knowledge deficits that can distort clinical reasoning \cite{ref65,ref66}. Historical choices should therefore be treated as evidence about decision structure, not ground truth.

AI can challenge proposed orders using broader evidence, consistency checks, and counterexamples. Unknown or previously unsuspected mechanisms remain a residual risk of any finite model. The framework does not claim to simulate hazards that are not scientifically plausible or observable under current evidence.

When new evidence appears, K, E, Q\_x, and M\_\{Q,E\} may all require revision. Purpose-conditioned abstraction is selective, not epistemically closed. Finally, this paper primarily addresses representation at a clinical decision point.

Longitudinal planning, repeated state transitions, and dynamic treatment regimes introduce additional questions about how ordered objectives evolve through time. State-abstraction and reinforcement-learning theory offer relevant machinery, but a full dynamic extension is beyond the present scope.

\section{Conclusion}\label{conclusion}

Clinical AI is often imagined as a progression from crude rules toward increasingly complete patient models and increasingly precise optimization. That trajectory can make simplification, thresholds, and stopping rules appear primitive - temporary compromises that better data and larger models will eventually remove. We argue for a different interpretation.

Every clinical model is an abstraction, and intelligence partly consists in choosing the right abstraction. Human planning research shows strategic simplification of task representations; clinical expertise shows selective problem representation, knowledge encapsulation, and goal-directed scripts. Model theory adds a second constraint: complexity that the available evidence cannot identify or calibrate may create detail without knowledge.

For purposive clinical AI, the objective is therefore more fundamental than it first appears. It does not merely rank outputs after a model has been built. It helps determine which distinctions deserve modeling and how finely they need to be represented.

Because clinical values are heterogeneous and comparative structure is often more defensible than global exchange rates, that objective should frequently be ordinal-first. Ordered threshold refinement then does double duty. It determines the decision and controls representational resolution.

The same dimension may recur at successively finer thresholds; the model is refined only when the decision reaches those thresholds. Once an earlier criterion leaves one admissible action, later distinctions do not need further computation for that choice. If a potentially decisive criterion remains unresolved, refinement is targeted to it.

The goal of better clinical AI is therefore not the most complete attainable virtual patient. It is the most defensible decision-sufficient abstraction of the patient for the task at hand: one that preserves what should matter first, models it only as finely as the evidence and decision require, and stops when action is justified. Purpose before fidelity.

Order before abstraction. Refine until decision-sufficient, not until reality-complete.

\appendix
\section{Minimal ordinally conditioned abstraction algorithm}\label{appendix-a.-minimal-ordinally-conditioned-abstraction-algorithm}

The following pseudocode is an implementation template rather than a claim of algorithmic novelty. It extends the ordered-threshold algorithm by making model resolution explicit.

\begin{Verbatim}[breaklines=true,fontsize=\small]
INPUT:
candidate actions A
clinical context x
broad domain knowledge K
available evidence E
ordered threshold program Q = (q1, q2, ..., qm)
S \<- A
for k = 1 to m:
determine what information/model resolution is needed to classify qk for actions in S
if current model is too coarse and refinement is evidence-supported:
refine only the relevant submodel / acquire targeted information
evaluate qk(a,x) for every a in S
if a potentially decisive qk remains unresolved:
pause ordered reduction at criterion k
obtain information, ask a preference question, seek expert review,
issue a conditional recommendation, or invoke an explicit act-now-versus-wait policy
do not let lower-priority criteria resolve the choice
if qk is resolved and at least one action passes and one fails:
S \<- {a in S : qk(a,x) = 1}
if \|S\| = 1:
return S
stop refining unreached lower-priority objectives for this decision
if multiple actions remain after all justified ordinal criteria:
return the surviving set, or apply a justified local cardinal refinement
LOG:
first discriminating threshold; provenance; evidence/model used;
refinements requested; unresolved states; criteria not reached; explicit revisions of Q
\end{Verbatim}

\section{Worked synthetic stress test for compensatory leakage}\label{appendix-b.-worked-synthetic-stress-test-for-compensatory-leakage}

Suppose treatments A and B tie on the first three criteria. At \(q_{4}\), severe irreversible toxicity \textless10\%, treatment A passes (8\%) and treatment B fails (14\%). Under the ordered program A wins; \(q_{5}\)-\(q_{10}\) are not reached.

\begin{longtable}[]{@{}
  >{\raggedright\arraybackslash}p{(\columnwidth - 8\tabcolsep) * \real{0.1400}}
  >{\raggedright\arraybackslash}p{(\columnwidth - 8\tabcolsep) * \real{0.3000}}
  >{\raggedright\arraybackslash}p{(\columnwidth - 8\tabcolsep) * \real{0.1800}}
  >{\raggedright\arraybackslash}p{(\columnwidth - 8\tabcolsep) * \real{0.1800}}
  >{\raggedright\arraybackslash}p{(\columnwidth - 8\tabcolsep) * \real{0.2000}}@{}}
\toprule\noalign{}
\begin{minipage}[b]{\linewidth}\raggedright
\textbf{Priority}
\end{minipage} & \begin{minipage}[b]{\linewidth}\raggedright
\textbf{Criterion}
\end{minipage} & \begin{minipage}[b]{\linewidth}\raggedright
\textbf{Treatment A}
\end{minipage} & \begin{minipage}[b]{\linewidth}\raggedright
\textbf{Treatment B}
\end{minipage} & \begin{minipage}[b]{\linewidth}\raggedright
\textbf{Ordered result}
\end{minipage} \\
\midrule\noalign{}
\endhead
\bottomrule\noalign{}
\endlastfoot
q1 & Mortality \textless10\% & 8\% pass & 9\% pass & continue \\
q2 & Paralysis \textless30\% & 25\% pass & 20\% pass & continue \\
q3 & Mortality \textless5\% & 8\% fail & 9\% fail & continue \\
q4 & Severe irreversible toxicity \textless10\% & 8\% pass & 14\% fail & A wins \\
q5 & Response \textgreater70\% & fail & pass & not reached \\
q6 & Oral dosing & fail & pass & not reached \\
q7 & \textless=1 visit/month & fail & pass & not reached \\
q8 & No routine laboratory monitoring & fail & pass & not reached \\
q9 & Price below preferred threshold & fail & pass & not reached \\
q10 & Early symptom relief & fail & pass & not reached \\
\end{longtable}

Now consider a finite scalar score after the \(q_{1}\)-\(q_{3}\) tie:

\[S(a) = 5q_{4}(a) + \sum_{k = 5}^{10}q_{k}(a).\]

Treatment A receives 5 while B receives 6 because six lower-priority wins collectively outweigh the finite coefficient attached to \(q_{4}\). The scalar system therefore selects B even though the specified program says \(q_{4}\) should discriminate before \(q_{5}\)-\(q_{10}\). This is compensatory leakage. A constrained or lexicographic scalar implementation could preserve the order; the example shows that assigning a large finite weight is not semantically equivalent to specifying priority.

The abstraction argument adds a second observation. Because \(q_{5}\)-\(q_{10}\) are never reached once \(q_{4}\) resolves the pair, a task-specific model does not need to increase \textbf{decision-directed} resolution solely to distinguish A and B on those later criteria. Predicting them may still be necessary for estimating an earlier objective, or may have scientific, safety-monitoring, or future-decision value; the claim is only that further resolution of those unreached objectives does not improve this choice under the stated program.
\section*{Declaration of generative AI and AI-assisted technologies in the manuscript preparation process}
During preparation of this work, the authors used OpenAI ChatGPT to support structural drafting, language refinement, literature organization, and manuscript formatting. The authors reviewed and edited the generated content as needed and take full responsibility for the content of the publication.

\end{document}